\documentclass[10pt,twocolumn,letterpaper]{article}

\usepackage{cvpr}              % To produce the CAMERA-READY version

\usepackage{booktabs}

\usepackage{graphicx}
\usepackage{amsmath}
\usepackage{amssymb}
\usepackage{booktabs}
\usepackage{multirow}
\usepackage{pifont}

\usepackage[accsupp]{axessibility}  % Improves PDF readability for those with disabilities.
\newcommand{\mainresultstable}{%
\begin{table*}[ht]
    \centering
    \caption{Main results on nuScenes domain generalization splits. We report mAP (\%) and NDS (\%) on the source domain (Src) and three target domains (Rain, Night, Boston). Average is computed over Rain/Night/Boston. L and C denote LiDAR and Camera, respectively. $\dagger$ represents the model trained with nuScenes full split.}
    \label{tab:main_results}
    \setlength{\tabcolsep}{10pt}
    \resizebox{0.95\linewidth}{!}{%
        \begin{tabular}{ccccccccccccc}
            \toprule
            \multirow{2}{*}[-0.6ex]{Model} & \multirow{2}{*}[-0.6ex]{Reference} & \multirow{2}{*}[-0.6ex]{Modality} & \multicolumn{2}{c}{Source} & \multicolumn{2}{c}{Rain} & \multicolumn{2}{c}{Night} & \multicolumn{2}{c}{Boston} & \multicolumn{2}{c}{Average} \\
            \cmidrule(lr){4-5} \cmidrule(lr){6-7} \cmidrule(lr){8-9} \cmidrule(lr){10-11} \cmidrule(lr){12-13}
            & & & mAP & NDS & mAP & NDS & mAP & NDS & mAP & NDS & mAP & NDS \\
            \midrule
            FSDv2~\cite{liFullySparseFusion2024}                 & TPAMI 24  & L    & 59.6   & 62.7   & 23.4    & 41.6    & 36.6     & 42.8     & 28.2      &  45.1  & 29.4 & 43.2 \\
            CMT~\cite{yanCrossModalTransformer2023}              & ICCV 23   & L+C   & 61.4   & 62.3   & 35.7    & 46.5    & 37.8     & 42.0     & 42.1      & 50.8      & 38.5 & 46.4 \\
            MOAD~\cite{chaRobustMultimodal3D2024}                & arXiv 24    & L+C   & 64.1   & 64.4   & 39.2    & 49.4    & 41.1     & 44.3     & 43.9      & 53.0      & 41.4 & 48.9 \\
            MEFormer~\cite{chaRobustMultimodal3D2024}            & arXiv 24      & L+C   & 63.4   & 63.7   & 40.1    & 50.1    & 41.1     & 44.0     & 44.3      & 53.0      & 41.8 & 49.0 \\
            ISFusion~\cite{yin2024isfusion}                      & CVPR 24   & L+C   & 66.3   & 65.4   & 39.8    & 49.4    & 41.8     & 45.1     & 45.4      & 53.4      & 42.3 & 49.3 \\
            MoME~\cite{mome}                                     & CVPR 25   & L+C   & 63.6   & 64.0   & 37.7    & 48.5    & 39.5     & 43.3     & 42.9      & 52.4      & 40.0 & 48.1 \\
            \midrule
            {Our baseline}                               & -         & L+C   & \textbf{68.4}   & 65.7   & 41.9    & 50.0    & 42.9     & 44.5     & 47.4      & 53.6      & 44.1 & 49.4 \\
            \textbf{CCF (Ours)}                                        &  -         & L+C   & 68.2   & \textbf{65.9}   & \textbf{44.7}    & \textbf{52.5}    & \textbf{44.2}     & \textbf{45.3}     & \textbf{50.6}      & \textbf{56.8}      & \textbf{46.5} & \textbf{51.5} \\
            \midrule
            CCF (Oracle){$^\dagger$}                                    &  -         & L+C   & 73.6   & 74.2   & 72.9    & 74.5    & 46.9     & 48.3     & 73.6     & 74.8 & 64.5 & 65.9 \\
            \bottomrule
        \end{tabular}%
    }
\end{table*}
}

\newcommand{\oracleresultstable}{%
\begin{table}[ht]
    \centering
    \caption{Oracle results on nuScenes domain generalization splits.}
    \label{tab:oracle_results}
    \setlength{\tabcolsep}{7pt}
    \resizebox{0.98\linewidth}{!}{%
    \begin{tabular}{ccccccc}
        \toprule
        \multirow{2}{*}[-0.6ex]{Models} & \multicolumn{2}{c}{All} & \multicolumn{2}{c}{Rain} & \multicolumn{2}{c}{Night}\\
        \cmidrule(lr){2-3} \cmidrule(lr){4-5} \cmidrule(lr){6-7}
        & mAP & NDS & mAP & NDS & mAP & NDS \\    \midrule
        BEVDet~\cite{huangBEVDetHighperformanceMulticamera2022}        & 34.9 & 41.7 & 30.1    & 43.2    & 12.0     & 21.7    \\
        PETR~\cite{PETR}                                               & 37.0 & 44.2 & 41.9    & 50.6    & 17.2     & 24.2    \\
        CenterPoint~\cite{centerpoint}                                 & 59.6 & 66.8   & 59.2    & -       & 35.4     & -       \\
        FSDv2~\cite{liFullySparseFusion2024}                           & 64.7 & 70.4   & 61.1    & 68.8    & 37.1     & 43.6    \\
        Transfusion-L~\cite{Transfusion}                               & 65.1 & 70.1   & 64.0    & 69.9    & 37.5     & 43.5    \\
        Transfusion~\cite{Transfusion}                                 & 67.3 & 71.2   & 67.5    & 71.9    & 39.8     & 44.7    \\
        BEVFusion~\cite{liu2022bevfusion}                              & 68.5 & 71.4   & 69.9    & -       & 42.8     & -       \\
        GraphBEV~\cite{GraphBEV}                                       & 70.1 & 72.9   & 70.2    & -       & 45.1     & -       \\
        CMT~\cite{yanCrossModalTransformer2023}                        & 70.3 & 72.9   & 70.5    & 73.7    & 42.8     & 46.3    \\  
        MOAD~\cite{chaRobustMultimodal3D2024}                          & 71.3 & 73.7  & 72.3    & 74.6    & 42.8     & 46.5    \\
        MEFormer~\cite{chaRobustMultimodal3D2024}                      & 71.5 & 73.9   & 72.3    & \textbf{75.0}    & 43.5     & 46.9    \\
        ISFusion~\cite{yin2024isfusion}                                & 72.7 & 73.8   & 72.8    & 74.8    & 45.0     & 47.4    \\       
        MoME~\cite{mome}                                               & 71.2 & 73.6   & 72.1    & 74.6    & 42.7     & 46.5    \\ 
        MV2DFusion~\cite{wangMV2DFusionLeveragingModalitySpecific2024} & 70.2 & 72.7   & 69.8    & 73.4    & 44.9     & 47.2    \\ % ! Single Frame
        \hline
        \textbf{Ours}                                                  & \textbf{73.6} & \textbf{74.2}  & \textbf{72.9}    & 74.5    & \textbf{46.9}     & \textbf{48.3}    \\ 
        \bottomrule
    \end{tabular}%
    }
\end{table}
}

\newcommand{\ablationtable}{%
\begin{table}[ht]
    \centering
    \caption{Ablation studies of the proposed components. We evaluate Query Decoupled Loss (DL), LiDAR-Guided Depth Prior (DP), and Complementary Cross-Modal Masking (CM), and report mAP (\%) and NDS (\%) on all domains.}
    \label{tab:ablation}
    \setlength{\tabcolsep}{4.5pt}
    \resizebox{0.98\linewidth}{!}{%
    \begin{tabular}{ccccccccccc}
        \toprule
        \multicolumn{3}{c}{Components} & \multicolumn{2}{c}{Source} & \multicolumn{2}{c}{Rain} & \multicolumn{2}{c}{Night} & \multicolumn{2}{c}{Boston} \\
        \cmidrule(lr){4-5} \cmidrule(lr){6-7} \cmidrule(lr){8-9} \cmidrule(lr){10-11}
        DL & DP & CM & mAP  & NDS  & mAP  & NDS  & mAP  & NDS  & mAP  & NDS  \\
        \midrule
        \ding{55}  & \ding{55} & \ding{55} & 68.4 & 65.7 & 41.9 & 50.0 & 42.9 & 44.5 & 47.4 & 53.6 \\
        \ding{55}  & \checkmark & \ding{55} & 68.6 & 66.0 & 42.3 & 50.6 & 41.8 & 44.4 & 47.5 & 53.8 \\
        \ding{55}  & \ding{55} & \checkmark & 68.4 & 65.8 & 44.5 & 51.4 & 43.4 & 45.2 & 49.6 & {56.4} \\
        \ding{55}  & \checkmark & \checkmark & 68.7 & 66.1 & \textbf{44.7} & 51.5 & 44.1 & 45.4 & 49.7 & 56.3 \\
        \midrule
        \checkmark & \ding{55} & \ding{55} & 68.4 & 66.1 & 42.8 & 52.4 & 42.1 & 44.7 & 48.1 & 56.0 \\
        \checkmark & \checkmark & \ding{55} & \textbf{68.8} & \textbf{66.4} & \textbf{44.7} & \textbf{53.6} & 42.2 & 44.6 & 50.0 & \textbf{56.9} \\
        \checkmark & \ding{55} & \checkmark & 68.4 & 66.2 & 44.3 & 51.2 & 43.9 & \textbf{45.8} & 50.2 & \textbf{56.9} \\
        \checkmark & \checkmark & \checkmark & 68.2 & 65.9 & \textbf{44.7} & 52.5 & \textbf{44.2} & 45.3 & \textbf{50.6} & 56.8 \\
        \bottomrule
    \end{tabular}%
    }
\end{table}
}

\newcommand{\gridmasktable}{%
\begin{table}[ht]
    \centering
    \caption{Ablation studies on Complementary Cross-Modal Mask variants. ``Cur.'' indicates whether curriculum learning is applied.}
    \label{tab:gridmask}
    \resizebox{0.47\textwidth}{!}{%
    \begin{tabular}{ccccccccccc}
        \toprule
        \multicolumn{3}{c}{Configuration} & \multicolumn{2}{c}{Rain} & \multicolumn{2}{c}{Night} & \multicolumn{2}{c}{Boston} \\
        \cmidrule(lr){1-3} \cmidrule(lr){4-5} \cmidrule(lr){6-7} \cmidrule(lr){8-9} \cmidrule(lr){10-11}
        & Variant & Cur. & mAP & NDS & mAP & NDS & mAP & NDS \\
        \midrule
        (a) & Image GridMask          & \ding{55}  & 42.8 & 50.7 & 42.5 & 45.3 & 48.3 & 54.7 \\
        \midrule
        (b) & Modal Mask              & \ding{55}  & 42.9 & 49.7 & 42.6 & 44.5 & 48.4 & 55.4 \\
        (c) & Consistent GridMask     & \ding{55}  & 42.8 & 49.8 & 42.2 & 44.3 & 48.9 & 55.5 \\
        (d) & Complementary GridMask   & \ding{55}  & 43.9 & 50.0 & 43.1 & 44.2 & 49.2 & 55.1 \\ % HPC6
        \midrule                 
        (e) & Complementary RandomMask & \checkmark & 44.1 & 51.1 & \textbf{44.1} & \textbf{46.0} & 49.7 & 55.2 \\
        (f) & Complementary GridMask   & \checkmark & \textbf{44.3} & \textbf{51.2} & 43.9 & 45.8 & \textbf{50.2} & \textbf{56.9} \\
        \bottomrule
    \end{tabular}%
    }
\end{table}
}

\definecolor{cvprblue}{rgb}{0.21,0.49,0.74}
\usepackage[pagebackref,breaklinks,colorlinks,allcolors=cvprblue]{hyperref}
\def\confName{CVPR}
\def\confYear{2026}

\title{CCF: Complementary Collaborative Fusion for Domain Generalized Multi-Modal 3D Object Detection}

\newcommand{\smallsection}[1]{\par\vspace{0.28em}\noindent\textbf{#1}}

\author{Yuchen Wu, Kun Wang, Yining Pan, Na Zhao\thanks{Corresponding Author}\\
Singapore University of Technology and Design\\
{\tt\small \{yuchen\_wu, yining\_pan\}@mymail.sutd.edu.sg, \{kun\_wang, na\_zhao\}@sutd.edu.sg}
}

\begin{document}
\maketitle
\begin{abstract}
Multi-modal fusion has emerged as a promising paradigm for accurate 3D object detection. However, performance degrades substantially when deployed in target domains different from training. In this work, focusing on dual-branch proposal-level detectors, we identify two factors that limit robust cross-domain generalization: 1) in challenging domains such as rain or nighttime, one modality may undergo severe degradation;
% -- a failure mode conspicuously absent from conventional training regimes dominated by daytime data; 
2) the LiDAR branch often dominates the detection process, leading to systematic underutilization of visual cues and vulnerability when point clouds are compromised. 
To address these challenges, we propose three components. 
First, Query-Decoupled Loss provides independent supervision for 2D-only, 3D-only, and fused queries, rebalancing gradient flow across modalities. 
Second, LiDAR-Guided Depth Prior augments 2D queries with instance-aware geometric priors through probabilistic fusion of image-predicted and LiDAR-derived depth distributions, improving their spatial initialization. 
Third, Complementary Cross-Modal Masking applies complementary spatial masks to the image and point cloud, encouraging queries from both modalities to compete within the fused decoder and thereby promoting adaptive fusion.
Extensive experiments demonstrate substantial gains over state-of-the-art baselines while preserving source-domain performance. Code and models are publicly available at \url{https://github.com/IMPL-Lab/CCF.git}.

\end{abstract}
\section{Introduction}
\label{sec:intro}

% =============================================
% Pilot Study Imbalance Evidence Placeholder Figure
% Replace placeholder boxes with actual plotted PDFs before camera-ready.
\begin{figure}\includegraphics[width=0.95\linewidth]{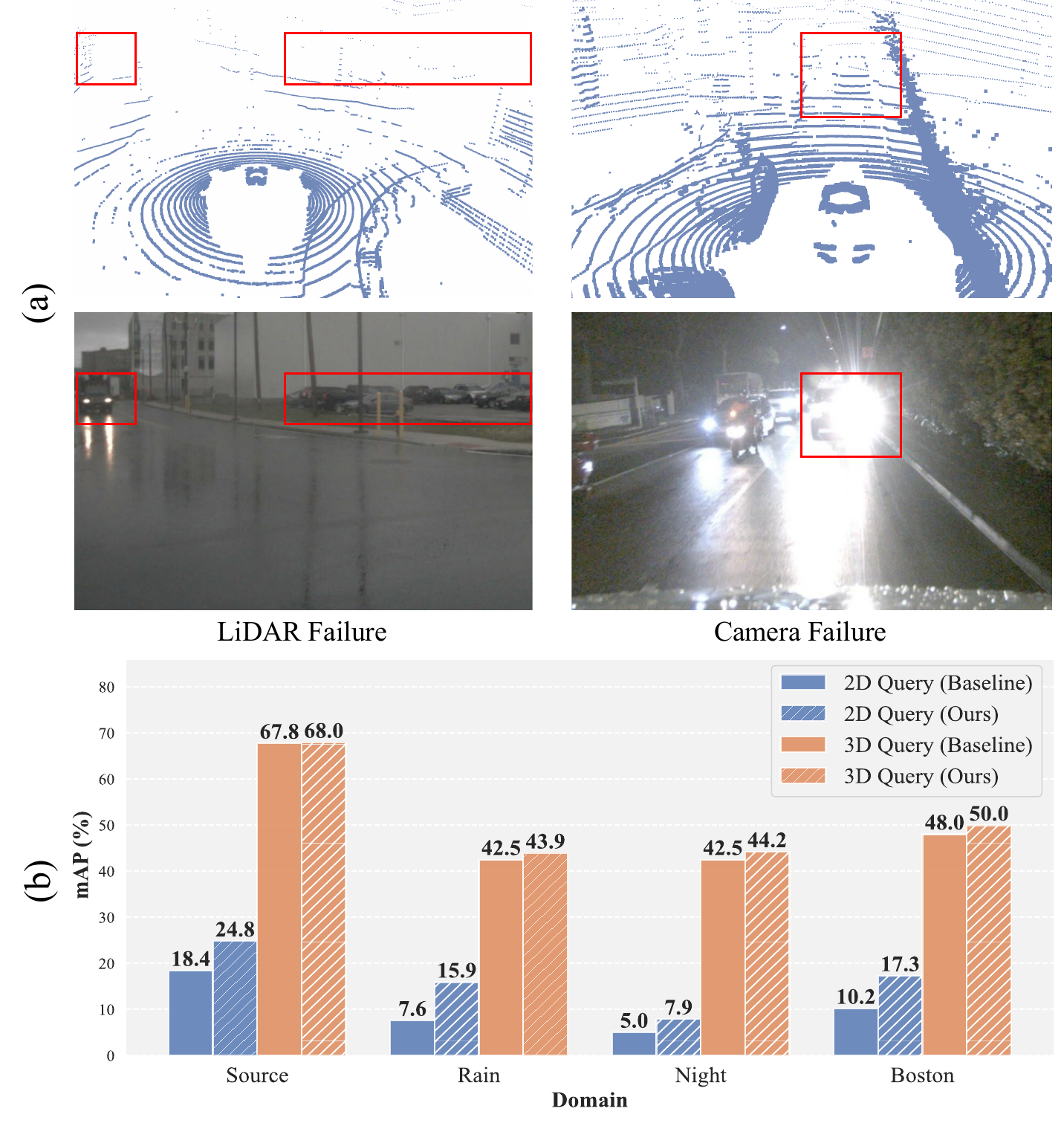}
	\caption{
    (a) Qualitative examples show that LiDAR and camera modalities degrade differently under adverse conditions. (b) Quantitative results on a baseline dual-branch detector show that our method substantially improves camera-originated queries and narrows their performance gap to LiDAR-originated queries.
    }
	\label{fig:introduction}
\end{figure}
% =============================================

3D scene understanding is a core capability for autonomous agents to perceive and interpret complex environments, and has been widely studied in tasks such as 3D object detection~\cite{centerpoint,zhao2020sess,sheng2022rethinking,han2024dual,Spgroup3d,Zhu_2025_CVPR,zhao2024sdcot++,sheng2025ct3d++,wang2025uncertainty}, 3D semantic segmentation~\cite{zhao2021ps2,zhao2021few,xu2023generalized,zhao2024synthetic}, and panoptic scene understanding~\cite{panopticph,IAL}. Among these tasks, multi-modal 3D object detection \cite{liu2022bevfusion,yanCrossModalTransformer2023,yin2024isfusion,wangMV2DFusionLeveragingModalitySpecific2024,liFullySparseFusion2024,xieSparseFusionFusingMultiModal2023,objfusion,cofix,bico,mv3d,paint} constitutes a pivotal paradigm for scene perception, leveraging the complementary strengths of LiDAR point clouds, which provide accurate geometric structure, and multi-view images, which offer rich semantic context. This heterogeneous modality integration has led to substantially improved detection performance on standard benchmarks. Nevertheless, existing methods largely focus on in-domain optimization and often suffer severe performance degradation when exposed to domain shifts caused by adverse weather, illumination changes, or unseen scene distributions. Given the diverse and dynamic nature of real-world environments, developing robust 3D object detection frameworks with strong cross-domain generalization is crucial for reliable deployment.

In this paper, we identify and address two key factors that contribute to cross-domain fragility. First, adverse weather and lighting conditions degrade sensor observations. As illustrated in Fig. \ref{fig:introduction} (a), heavy rain causes attenuation and scattering in LiDAR returns, resulting in sparse point clouds for the left and distant vehicles. Meanwhile, low visibility combined with strong glare introduces photometric distortions in camera images, making image-based localization more difficult. Such conditions are rarely represented in training datasets, which are dominated by daytime scenes, leaving models insufficiently prepared for cross-domain deployment. Second, we focus on dual-branch proposal-level detectors, where camera and LiDAR branches generate modality-specific queries before fusion, and observe that existing methods exhibit modality imbalance with reliance on the LiDAR branch. As shown in Fig. \ref{fig:introduction} (b), detection performance with 2D queries is consistently lower than 3D queries across all evaluated domains, indicating that the semantic information in images is not fully utilized. As a result, when LiDAR quality deteriorates, detection performance can degrade sharply, revealing a critical weakness in the robustness of current multi-modal frameworks.

To address the aforementioned challenges, we propose CCF (Complementary Collaborative Fusion), a systematic framework designed to mitigate modality imbalance and enhance cross-domain robustness. Our approach consists of three components. First, to mitigate supervision imbalance, \textbf{Query Decoupled Loss} provides independent learning pathways for 2D-only, 3D-only, and fused queries with three parallel, weight-shared decoder passes. This ensures that image-based queries receive dedicated gradient flow, preventing them from being overshadowed by their 3D counterparts. Second, to improve the geometric accuracy of 2D proposals, \textbf{LiDAR-Guided Depth Prior} adaptively fuses learned image-based depth estimates with direct geometric priors from LiDAR points, significantly enhancing 2D query initialization. Finally, to ensure these enhanced and well-supervised queries are effectively utilized during fusion, \textbf{Complementary Cross-Modal Masking} introduces a novel augmentation strategy that applies complementary spatial masks to both image and point cloud modalities. This design simulates localized sensor degradations commonly observed under adverse conditions and, more importantly, forces queries from both modalities to compete within the fused decoder, thereby promoting adaptive fusion and preventing over-reliance on any single modality.

Our contributions are summarized as follows:
\begin{itemize}[leftmargin=*,noitemsep,topsep=2pt]
    \item We identify and analyze the modality imbalance problem in dual-branch multi-modal 3D detectors, showing that camera-originated queries are systematically underutilized in supervision, initialization, and fusion.
    \item We propose CCF (Complementary Collaborative Fusion), a unified framework that improves balanced modality utilization through Query Decoupled Loss, LiDAR-Guided Depth Prior, and Complementary Cross-Modal Masking.
    \item Extensive experiments on a realistic nuScenes-based domain shift benchmark demonstrate state-of-the-art cross-domain performance and consistent robustness gains across diverse target domains.
\end{itemize}

\section{Related Works}

\subsection{Multi-Modal Fusion For 3D Object Detection}
Multi-modal 3D object detection methods can be broadly grouped by how object representations are formed. One line of work first builds a shared multi-modal representation and then performs decoding with unified object queries. BEVFusion~\cite{liu2022bevfusion} projects both modalities into a common BEV space, while TransFusion~\cite{Transfusion} and CMT~\cite{yanCrossModalTransformer2023} further perform object reasoning with transformer-based queries on fused features. In contrast, proposal-driven methods explicitly maintain modality-specific object hypotheses before fusion. F-PointNet~\cite{fpnet} uses 2D detections to guide 3D object search in point clouds, while MV2DFusion~\cite{wangMV2DFusionLeveragingModalitySpecific2024} generates 2D and 3D proposals from separate branches and refines them jointly with multi-modal features. This explicit dual-branch design provides a natural basis for studying the imbalance between camera- and LiDAR-originated queries.

\subsection{Robust Multi-Modal 3D Object Detection}
Although multi-modal detectors have achieved strong performance in standard benchmarks, improving their robustness remains crucial for real-world autonomous driving. Existing robust multi-modal 3D detection methods mainly focus on sensor corruption or missing-modality scenarios. MetaBEV~\cite{metabev2025} introduces a modality-arbitrary BEV decoder that updates meta-BEV queries from available sensors via cross-modal deformable attention. UniBEV~\cite{wang2023unibev} introduces channel-normalized weighted fusion for robust feature aggregation when one modality is unavailable. CMT~\cite{yanCrossModalTransformer2023} adopts sensor dropout as data augmentation to improve robustness against sensor failures. MEFormer~\cite{chaRobustMultimodal3D2024} proposes modality-agnostic decoding to reduce over-reliance on LiDAR, while MoME~\cite{mome} employs parallel expert decoders to decouple modality dependencies under sensor-failure settings. In contrast, our work addresses domain shifts induced by environmental changes, where both modalities remain available but exhibit different reliability. Such a setting requires not only robustness to degradation, but also balanced utilization of camera and LiDAR information, which is largely overlooked by prior methods.

\subsection{Data Augmentation for 3D Perception}
Data augmentation creates diverse training distributions that improve robustness to noise and bias in real-world environments. PolarMix~\cite{xiao2022polarmix} enriches point cloud distributions through cross-scan mixing, while LaserMix~\cite{kong2023lasermix} exchanges LiDAR beams across inclination ranges. Park et al.~\cite{park2024rethinking} learn erase patterns to mimic point drop under adverse weather. Unlike these methods, our augmentation is not designed to simply simulate corruption patterns, but to reshape the competition between camera- and LiDAR-based queries under complementary partial observations.
% =============================================
% SECTION 3: PILOT STUDY
% =============================================
\section{Pilot Study: Unveiling Modality Imbalance}
\label{sec:pilot}

% =============================================
% Pilot Study Figure Placeholder
% TODO: Create multi-panel figure showing:
% - Panel (a): 2D AP comparison (projected 3D queries vs 2D proposals vs 2D queries)
% - Panel (b): Hungarian matching selection statistics (3D vs 2D query counts)
% - Panel (c): Performance breakdown (2D-only, 3D-only, Fused queries) across domains
% =============================================
\begin{figure}
	\centering
	\includegraphics[width=0.85\linewidth]{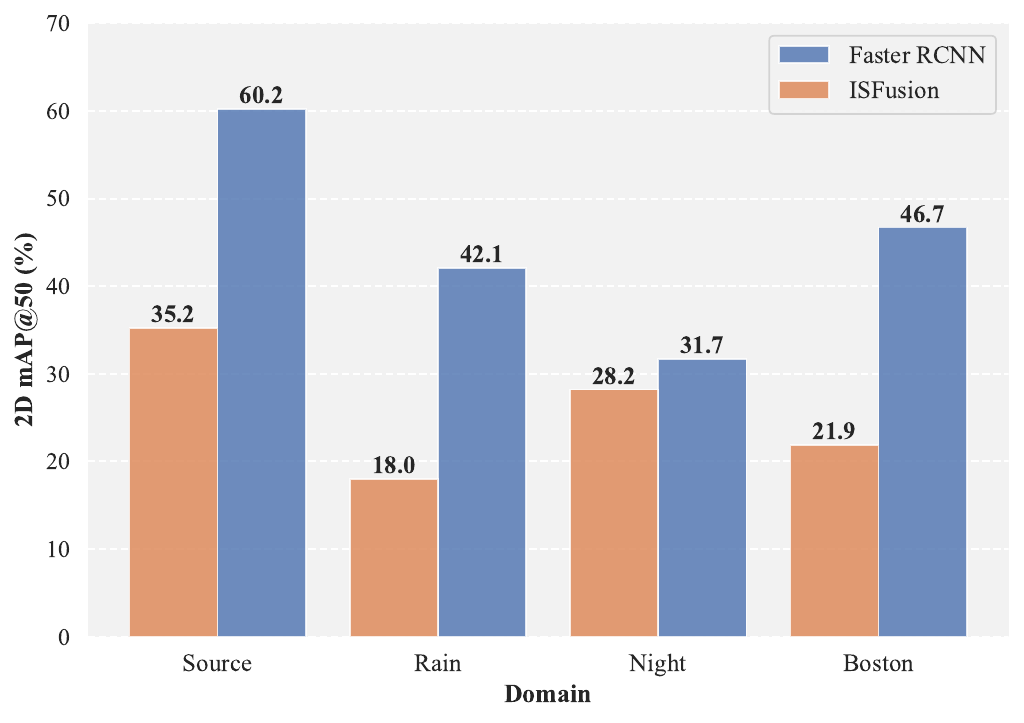}
	\caption{Analysis of 2D proposal quality. We compare the 2D mAP@50 of proposals from the 2D detector (Faster R-CNN) against projected 3D boxes from the 3D detector (ISFusion). The results show that native 2D proposals consistently outperform projected 3D proposals across all domains.}
	\label{fig:pilot_study}
\end{figure}
% =============================================

% Paragraph 1: Motivation and setup
Dual-branch detectors have demonstrated competitive performance by explicitly leveraging modality-specific proposals~\cite{wangMV2DFusionLeveragingModalitySpecific2024,xieSparseFusionFusingMultiModal2023}. However, their performance under domain shift remains underexplored. To investigate this, we conduct a pilot study on a representative framework: MV2DFusion~\cite{wangMV2DFusionLeveragingModalitySpecific2024} equipped with ISFusion~\cite{yin2024isfusion} as the 3D proposal generator and Faster R-CNN~\cite{fasterrcnn} as the 2D proposal generator, both trained on our source domain (clear daytime Singapore) and evaluated across all splits. Preliminary observation reveals a striking imbalance: when evaluated separately, 2D-originated queries achieve only 18.44\% 3D mAP on the source domain, while 3D-originated queries reach 67.75\% mAP. To explain this gap, we first analyze the 2D branch potential, and then examine two limiting factors: supervision imbalance and inaccurate depth estimation.
% —a significant performance gap. We investigate two critical aspects to understand this phenomenon: proposal quality and training supervision.

% Paragraph 2: Proposal quality analysis - 2D potential
\smallsection{Untapped 2D Proposal Quality.}
We first examine whether the quality of initial proposals accounts for the observed performance gap between 2D and 3D queries. To isolate proposal quality from downstream fusion effects, we project 3D boxes predicted by ISFusion onto the image plane and evaluate their 2D Average Precision (AP) against Faster R-CNN. As shown in \cref{fig:pilot_study}, image proposals maintain consistently high 2D AP across all domains, surpassing projected 3D boxes from ISFusion. This indicates that 2D proposals retain strong semantic quality even under domain shift.

% Paragraph 3: Training supervision analysis
\smallsection{Imbalanced Training Supervision.}
To better understand this underutilization, we examine the supervision allocation induced by Hungarian matching during training. Specifically, we count the matched 2D, 3D, and fused queries over 20 training epochs on the source domain. The resulting statistics reveal a pronounced supervision imbalance: each training sample produces, on average, 9.375 matched 3D queries but only 0.25 matched 2D queries, corresponding to a 37.5:1 ratio. This result shows that 3D queries overwhelmingly dominate the assignment process and therefore receive substantially stronger gradient supervision.

% Paragraph 4: Depth estimation quality

\smallsection{Inaccurate Depth Estimation.}
Beyond semantic quality and supervision imbalance, we further examine whether inaccurate depth estimation limits the effectiveness of 2D queries, since depth initialization is critical for 3D localization. In dual-branch frameworks~\cite{wangMV2DFusionLeveragingModalitySpecific2024,xieSparseFusionFusingMultiModal2023}, 2D queries estimate depth from image RoI features using a learned predictor, without explicit geometric constraints. To quantify this limitation, we measure the Mean Absolute Error (MAE) of matched 2D-query depth predictions against ground-truth 3D boxes over the 0--40 m range across all domains. The results reveal substantial errors even on the source domain, with an MAE of 1.78 m, and the error further increases under domain shift to 3.01 m on Rain, 2.27 m on Night, and 2.55 m on Boston. These findings indicate that purely image-based depth prediction introduces considerable localization uncertainty under challenging conditions. This observation also suggests a natural direction for improvement: leveraging LiDAR-derived geometric priors to provide more reliable depth cues for 2D queries.

% Paragraph 5: Summary of findings → transition to method
\smallsection{Key Findings.}
Our pilot study yields three key observations: (1) image proposals retain strong untapped potential, exhibiting high 2D detection quality across domains; (2) supervision is disproportionately allocated to 3D queries, leaving 2D queries insufficiently optimized; and (3) inaccurate depth estimation further limits the 3D localization capability of 2D queries. Together, these findings motivate the design of mechanisms that \textit{improve 2D query initialization} with geometric cues and \textit{rebalance supervision} during training. We introduce these components in \cref{sec:method}.

% =============================================
% SECTION 4: METHOD
% =============================================
\section{Method}
\label{sec:method}

% =============================================
% Overview paragraph introducing the problem and solution pipeline
% =============================================

In this section, we present CCF, a framework designed to mitigate the modality imbalance identified in \cref{sec:pilot}. We begin by revisiting the baseline dual-branch architecture (\cref{sec:method:baseline}) to establish necessary notation and context. We then introduce three components: \textbf{Query Decoupled Loss} (\cref{sec:method:decoupled_loss}) to provide balanced supervision across modality-specific queries, \textbf{LiDAR-Guided Depth Prior} (\cref{sec:method:depth_prior}) to enhance 2D query initialization with geometric cues, and \textbf{Complementary Cross-Modal Masking} (\cref{sec:method:masking}) to encourage complementary learning during fusion. 
% Together, these components enable balanced utilization of queries from both modalities under domain shift.

% =============================================
% Method Overview Figure - PLACEHOLDER
% TODO: Create comprehensive method overview figure showing:
% - Top: Overall pipeline with dual-branch architecture (2D/3D proposal generation → query-based fusion)
% - Component 1: Query Decoupled Loss (three parallel decoder passes with weight sharing)
% - Component 2: LiDAR-Guided Depth Prior (dual-source depth distributions with adaptive fusion)
% - Component 3: Inconsistent Cross-Modal Masking (inverse spatial masking on image and point cloud)
% - Visual flow showing how each component addresses specific limitation from pilot study
% =============================================
\begin{figure*}[t]
	\centering
	\includegraphics[width=\linewidth]{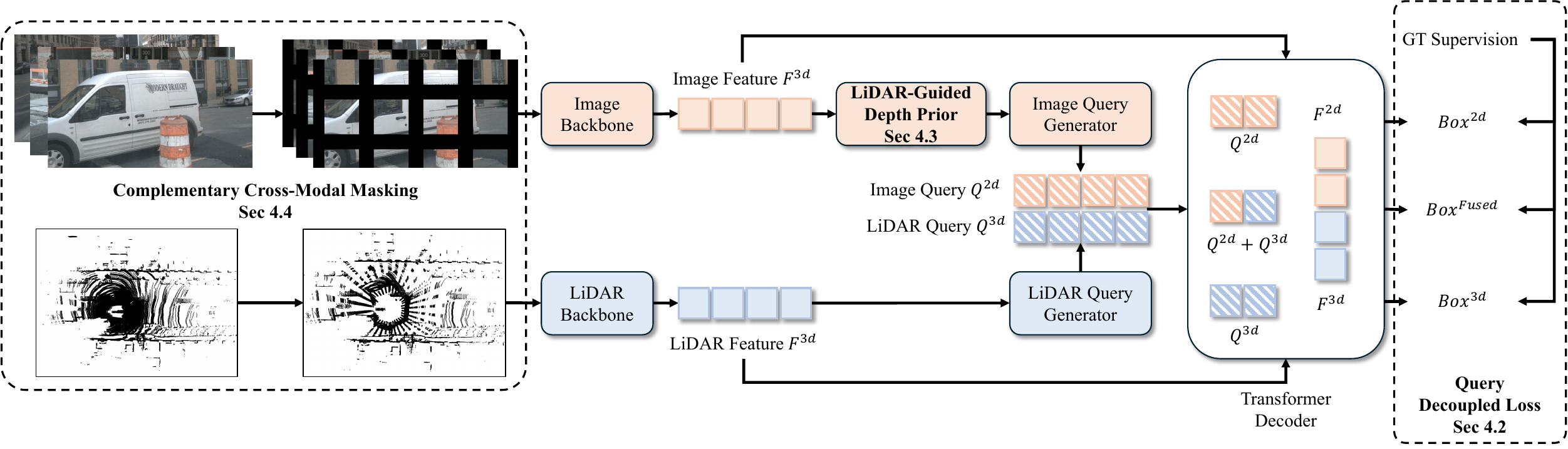}
	\caption{\textbf{Overview of CCF.} CCF addresses modality imbalance with three components. \textbf{(a) Query Decoupled Loss} uses three parallel, weight-shared decoder passes (2D-only, 3D-only, and fused) to provide modality-specific supervision while avoiding shortcut learning. \textbf{(b) LiDAR-Guided Depth Prior} adaptively fuses image-predicted and LiDAR-derived depth distributions to improve 2D query initialization. \textbf{(c) Complementary Cross-Modal Masking} applies complementary spatial masking, encouraging balanced competition between camera- and LiDAR-originated queries. Together, these components improve modality balance and robustness under domain shift.}
	\label{fig:method_overview}
\end{figure*}
% =============================================

% =============================================
% Subsection 4.1: Baseline Architecture Review
% =============================================
\subsection{Revisiting Dual-Branch Detection}
\label{sec:method:baseline}

% Paragraph 1: Overview
Our method builds upon the dual-branch detection framework MV2DFusion~\cite{wangMV2DFusionLeveragingModalitySpecific2024}, which follows a two-stage pipeline: (1) modality-specific proposal generation, and (2) query-based fusion. A 2D detector produces $M^{2d}$ 2D proposals $\mathbf{b}^{2d} \in \mathbb{R}^{M^{2d} \times 4}$, while a 3D detector generates $M^{3d}$ 3D proposals $\mathbf{b}^{3d} \in \mathbb{R}^{M^{3d} \times 7}$. These proposals are then converted into queries for fusion.

% Paragraph 2: Query formulation
\smallsection{Query Formulation.}
Each 3D proposal generates a query $\mathbf{q}^{3d} = (\mathbf{c}^{3d}, \mathbf{r}^{3d})$, where $\mathbf{c}^{3d} \in \mathbb{R}^{M^{3d} \times C}$ contains RoI appearance and geometric features, and $\mathbf{r}^{3d} \in \mathbb{R}^{M^{3d} \times 3}$ takes the centers of $\mathbf{b}^{3d}$ as reference points. Each 2D proposal produces a query $\mathbf{q}^{2d} = (\mathbf{c}^{2d}, \mathbf{r}^{2d})$, where $\mathbf{c}^{2d} \in \mathbb{R}^{M^{2d} \times C}$ encodes RoI features, and $\mathbf{r}^{2d} \in \mathbb{R}^{M^{2d} \times 3}$ represents estimated 3D positions derived from depth prediction.

% Paragraph 3: Decoder and training
\smallsection{Decoder and Training.}
Combined queries $\mathbf{q}^0 = (\mathbf{q}^{2d}, \mathbf{q}^{3d})$ are processed through a transformer decoder with $L$ layers, producing refined queries $\mathbf{q}^L$ for final 3D box prediction. Training employs Hungarian matching to assign queries to ground truth. As revealed in \cref{sec:pilot}, this standard training leads to severe imbalance where 3D queries dominate matching, leaving 2D queries insufficiently supervised.

% =============================================
% Subsection 3.2: Query Decoupled Loss
% =============================================
\subsection{Query Decoupled Loss}
\label{sec:method:decoupled_loss}

% Paragraph 1: Motivation
As revealed in \cref{sec:pilot}, 3D queries dominate Hungarian matching during training and therefore receive substantially more supervision than 2D queries. This imbalance limits the gradient signal reaching the image branch and weakens its optimization. Although both query types are matched under the same assignment rule, the superior localization quality of 3D queries, inherited from LiDAR geometry, allows them to capture most ground-truth assignments. As a result, 2D queries receive insufficient supervision to improve their own localization quality, further reinforcing the imbalance. To mitigate this issue, we propose Query Decoupled Loss, which provides independent supervision for each modality and thereby rebalances the training process.

% Paragraph 2: Method design
\smallsection{Decoupled Decoder Architecture.}
A straightforward alternative is to decode the fused queries $\mathbf{q}^L$ in a single pass and then separate them by modality for independent loss computation. However, this design introduces a shortcut: during self-attention, 2D queries can rely on information propagated from co-attending 3D queries, rather than being optimized as an independent query set. To avoid this issue, we execute the decoder \textit{three times in parallel} with shared weights: (1) a \textit{2D-only pass} operating on $\mathbf{q}^{2d,0}$, (2) a \textit{3D-only pass} operating on $\mathbf{q}^{3d,0}$, and (3) a \textit{fused pass} operating on the concatenated query set $\mathbf{q}^0 = (\mathbf{q}^{2d,0}, \mathbf{q}^{3d,0})$. All three passes attend to the same multi-modal feature tokens, while maintaining separate query sets in self-attention. During inference, only the fused pass is used for prediction, introducing no additional computational cost.

% Paragraph 3: Mathematical formulation
\smallsection{Loss Formulation.}
Each decoder pass produces refined queries: $\mathbf{q}^{2d,L}$ from the 2D-only pass, $\mathbf{q}^{3d,L}$ from the 3D-only pass, and $\mathbf{q}^L$ from the fused pass. Independent Hungarian matching and losses are applied to each output:
\begin{equation}
\mathcal{L}_{\text{total}} = \mathcal{L}_{2d} + \mathcal{L}_{3d} + \mathcal{L}_{\text{fused}},
\label{eq:decoupled_loss}
\end{equation}
where each component consists of classification and box regression terms:
\begin{equation}
\mathcal{L}_{(\cdot)} = \mathcal{L}_{\text{cls}}^{(\cdot)} + \mathcal{L}_{\text{box}}^{(\cdot)}.
\label{eq:branch_loss}
\end{equation}
We adopt focal loss~\cite{focalloss} for $\mathcal{L}_{\text{cls}}$ and L1 loss for $\mathcal{L}_{\text{box}}$, following the baseline. Importantly, the 2D-only branch ensures that 2D queries receive direct supervision by competing only within the 2D query set, while the fused branch preserves the full capability of the original framework.

% % Paragraph 4: Comparison with related work
% \smallsection{Relation to Prior Work.}
% Our approach differs fundamentally from MOAD's Token Decouple Loss~\cite{cite_mome}. MOAD addresses sensor failure scenarios by training separate decoders for each modality, enabling test-time robustness when one sensor is missing. In contrast, we assume both sensors are always present but face a supervision imbalance problem during training. Our weight-shared decoder with parallel passes preserves cross-modal learning in the fused branch while preventing shortcut learning through isolated 2D/3D branches. This design targets balanced utilization under domain shift rather than sensor dropout robustness.

% =============================================
% Subsection 3.3: LiDAR-Guided Depth Prior
% =============================================
\subsection{LiDAR-Guided Depth Prior}
\label{sec:method:depth_prior}

\begin{figure}
	\centering
	\includegraphics[width=\linewidth]{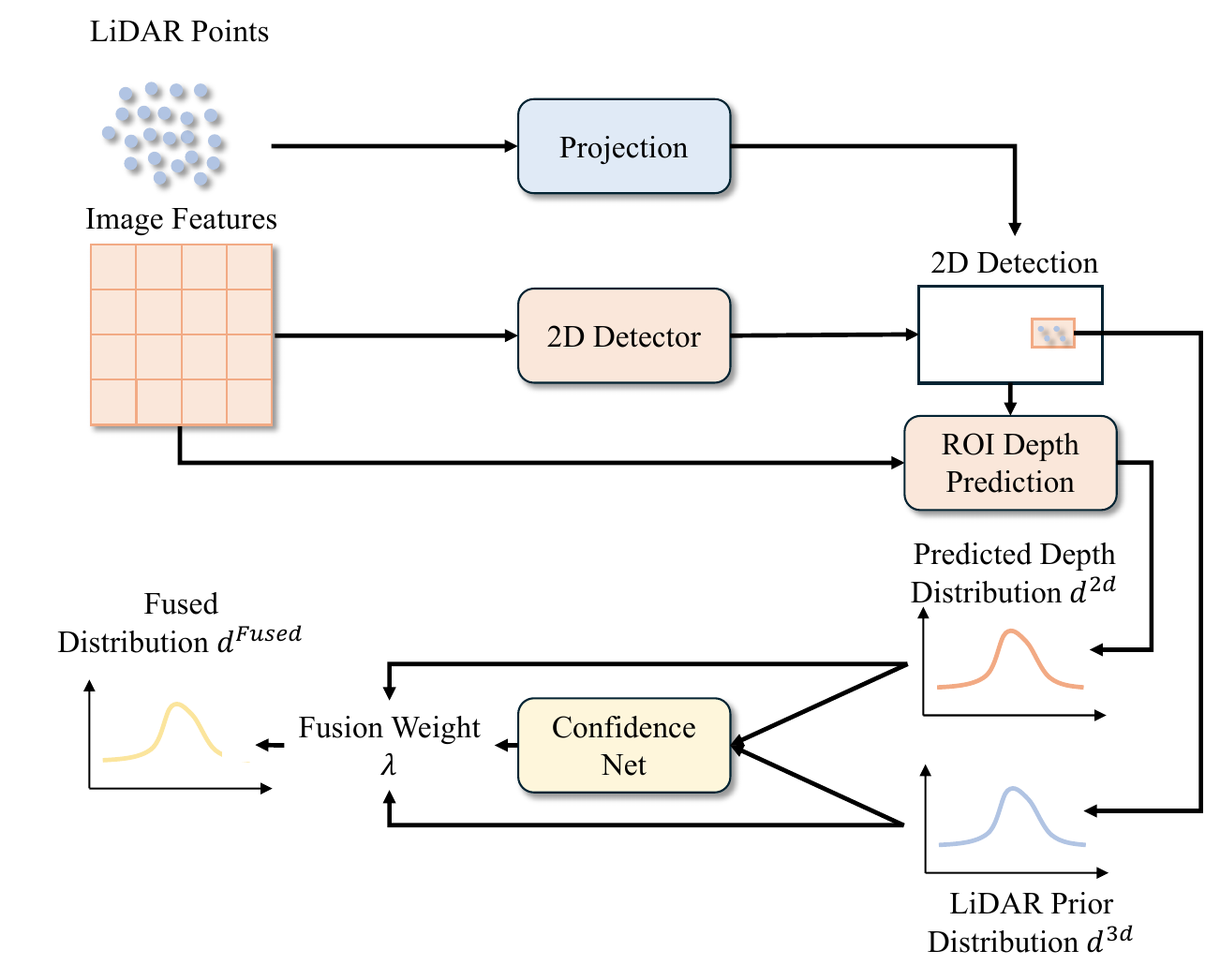}
	\caption{\textbf{Illustration of our LiDAR-Guided Depth Prior.} For each 2D proposal, we extract a learned depth distribution from image features ($\mathbf{d}^{2d}$) and a geometric prior from LiDAR points ($\mathbf{d}^{3d}$). A confidence network adaptively predicts a fusion weight ($\lambda$) to combine them into a fused distribution ($\mathbf{d}^{Fused}$), which provides robust depth initialization for the 2D query.}
	\label{fig:depth_prior}
\end{figure}

% Paragraph 1: Motivation
While Query Decoupled Loss improves the supervision received by 2D queries in isolation, it does not address the fundamental quality issue identified in \cref{sec:pilot}: inaccurate depth estimation still severely limits their 3D localization capability. Consequently, even under balanced supervision, poorly initialized 2D queries remain less competitive than 3D queries in the fused branch. In multi-modal detection, however, LiDAR points within 2D proposals can provide informative estimates of the underlying object depth distribution. Motivated by this observation, we leverage LiDAR-derived geometric priors to enhance 2D query initialization and improve their effectiveness during fusion.

% Paragraph 2: Depth distribution extraction
\smallsection{Dual-Source Depth Distributions.}
As illustrated in \cref{fig:depth_prior}, for each 2D proposal $\mathbf{b}^{2d}_i$, we estimate depth from two complementary sources. From the image branch, a lightweight depth predictor processes RoI features and outputs a probability distribution $\mathbf{d}^{2d}_i \in \mathbb{R}^{D}$ over $D$ depth bins. From the LiDAR branch, we collect points inside the 3D frustum defined by $\mathbf{b}^{2d}_i$ and discretize their depths into a histogram. If no LiDAR points fall inside the frustum, we use a uniform distribution by default. The normalized histogram forms a geometric prior distribution $\mathbf{d}^{3d}_i \in \mathbb{R}^{D}$. This LiDAR-derived prior provides explicit geometric evidence, although it may become sparse or noisy for distant objects or under adverse conditions such as rain.

% Paragraph 3: Fusion mechanism
\smallsection{Adaptive Distribution Fusion.}
Simply averaging or concatenating these two distributions is suboptimal, since their reliability varies across instances. For example, image-based depth may be more reliable for distant objects with sparse LiDAR support, whereas the LiDAR prior is often more accurate for nearby objects with sufficient point coverage. To address this issue, we introduce a confidence network that adaptively fuses these complementary distributions. Given $\mathbf{d}^{2d}_i$ and $\mathbf{d}^{3d}_i$, this lightweight network predicts an instance-specific fusion weight $\lambda_i \in [0,1]$. The fused depth distribution is then computed as
\begin{equation}
\mathbf{d}^{fused}_i = \sigma\left(\lambda_i \cdot \log(\mathbf{d}^{2d}_i) + (1-\lambda_i) \cdot \log(\mathbf{d}^{3d}_i)\right),
\label{eq:depth_fusion}
\end{equation}
where $\sigma(\cdot)$ denotes the softmax function. This log-space fusion, analogous to a Product-of-Experts~\cite{poe}, allows the model to emphasize the more reliable depth source for each instance. We then compute the expected depth from $\mathbf{d}^{fused}_i$ and use it to initialize the 3D reference point $\mathbf{r}^{2d}_i$ of the corresponding 2D query, thereby improving its localization.

% % Paragraph 4: Instance-wise design advantages
% \smallsection{Instance-Level Adaptation.}
% A critical advantage of our design is its instance-level adaptivity. Unlike image-level geometry transfer methods such as SparseFusion~\cite{cite_sparsefusion}, which apply global depth regularization, our confidence network learns to weight modalities differently for each object based on local geometric reliability. For example, distant objects with sparse LiDAR points may rely more on image-based estimation ($w^{img}_i > w^{pc}_i$), while nearby objects with dense point coverage benefit from geometric priors ($w^{pc}_i > w^{img}_i$). This instance-wise adaptation is particularly suited for dual-branch architectures where proposals explicitly represent individual objects, and proves robust across domain shifts where modality reliability varies spatially within each scene.

% =============================================
% Subsection 3.4: Inconsistent Cross-Modal Masking
% =============================================
\subsection{Complementary Cross-Modal Masking}
\label{sec:method:masking}

% Paragraph 1: Motivation
While the previous two components strengthen 2D queries through dedicated supervision and improved initialization, a critical challenge remains in the fused branch: 3D queries still dominate Hungarian matching in the fused branch. The decoder learns to rely heavily on consistently accurate 3D queries, limiting its ability to adaptively leverage 2D queries when conditions favor them. To address this, we need an augmentation strategy that forces the decoder to utilize queries from both modalities during training, preparing it to make adaptive selections at test time based on domain-specific modality reliability.

% Paragraph 2: Core design with implementation
\smallsection{Complementary Cross-Modal Masking.}
% \paragraph{Complementary Cross-Modal Masking.}
In real-world domain shifts, modalities often degrade differently: rain obscures LiDAR while cameras remain informative, whereas low-light conditions degrade images while LiDAR maintains geometric coverage. This complementary degradation pattern suggests that two modalities should be augmented complementarily rather than uniformly during training. We simulate this through complementary spatial masking at the input level. Given a spatial mask $\mathbf{M} \in \{0,1\}^{H \times W}$ on the image plane, we mask the raw image $\mathbf{I}$ by setting pixels with $\mathbf{M}=0$ to zero. For the LiDAR point cloud $\mathbf{P}=\{\mathbf{p}_i\}_{i=1}^N$, we project each point onto the image plane via camera intrinsics $\mathbf{K}$ and extrinsics $[\mathbf{R}|\mathbf{t}]$:
% \begin{equation}
% [u_i, v_i, 1]^\top \propto \mathbf{K}[\mathbf{R}|\mathbf{t}]\mathbf{p}_i.
% \label{eq:projection}
% \end{equation}
\begin{equation}
	[u_i, v_i, 1]^\top \propto \mathbf{K}(\mathbf{R}\mathbf{p}_i + \mathbf{t}),
	\label{eq:projection}
\end{equation}
and retain only the points whose projected locations fall in the complementary masked regions. In this way, when one modality is masked at a spatial location, the other modality is explicitly retained. Unlike consistent masking that degrades both modalities simultaneously, our design encourages more balanced competition between camera- and LiDAR-originated queries under partial observations.

% We then apply the {inverse} mask, retaining only points where $\mathbf{M}[u_i, v_i]=0$ and dropping those where $\mathbf{M}[u_i, v_i]=1$. This creates spatially-aligned but informationally-complementary patterns: when image pixels are masked, the corresponding LiDAR points are preserved, and vice versa. Unlike consistent masking that degrades both modalities simultaneously, our approach preserves complementary observations across modalities at each masked region. 
For mask generation, we adopt GridMask~\cite{gridmask}, following the 2D augmentation protocol used in CMT~\cite{yanCrossModalTransformer2023}, to produce structured spatial masks. To stabilize training, we use a curriculum schedule in which the masking probability increases linearly from 0 to $p$ during training. This allows the model to first learn from complete multi-modal observations before progressively adapting to partial inputs. The masked images and filtered point clouds are then fed into their respective feature extractors prior to query generation.

% Paragraph 3: Benefits for domain generalization
\smallsection{Benefits for Adaptive Query Selection.}
This strategy better reflects real-world domain shifts in which the two modalities degrade asymmetrically, such as LiDAR sparsity in rain and visual degradation at night. By training with complementary partial observations, the decoder learns to adaptively select and balance camera- and LiDAR-originated queries according to their reliability. Unlike complete modality dropout in CMT, our approach preserves both modalities with complementary visibility patterns, better matching practical test conditions where both sensors remain available but differ in quality.

% =============================================
% Subsection 3.5: Training and Inference (Optional)
% =============================================
% \subsection{Training and Inference}
% \label{sec:method:training}

% % Paragraph 1: Training strategy
% % - Apply all three components jointly
% % - Augmentation schedule for cross-modal masking
% % - Loss balancing across decoupled branches

% % Paragraph 2: Inference
% % - Use fused queries for final predictions
% % - No additional computational overhead compared to baseline
% % - Adaptive modality utilization learned during training
% =============================================
% SECTION 4: EXPERIMENTS
% =============================================
\section{Experiments}
\label{sec:experiments}

% =============================================
% Subsection 4.1: Experimental Setup
% =============================================
\subsection{Experimental Setup}
\mainresultstable
\label{sec:exp:setup}

% Paragraph 1: Dataset and Domain Splits
\smallsection{Dataset and Domain Splits.}
We build our benchmark on nuScenes~\cite{Caesar2019nuScenesAM} by defining domain splits according to natural environmental attributes.
The \textbf{source domain} consists of 226 clear daytime Singapore scenes. We consider three \textbf{target domains} from the validation set: \textit{Rain} (27 scenes), \textit{Night} (15 scenes), and \textit{Boston} (77 scenes), covering naturally occurring weather, illumination, and geographic shift.

% Paragraph 2: Implementation Details
\smallsection{Implementation Details.}
We implement CCF on top of the MV2DFusion architecture~\cite{wangMV2DFusionLeveragingModalitySpecific2024}, using ISFusion~\cite{yin2024isfusion} and Faster R-CNN~\cite{fasterrcnn} as the LiDAR and image query generators, respectively, followed by a 6-layer fusion decoder. Query Decoupled Loss uses identical weights across the 2D-only, 3D-only, and fused branches. LiDAR-Guided Depth Prior employs $D=25$ bins with a 3-layer MLP confidence network. Complementary Cross-Modal Masking adopts GridMask~\cite{gridmask} with curriculum learning, increasing masking probability from 0 to $p=0.7$.

\smallsection{Training Procedure.}
To prevent data leakage and ensure fair evaluation, we adopt a two-stage training strategy. In \textit{Stage 1}, both proposal generators are pre-trained exclusively on the source domain. Specifically, the 2D detector (Faster R-CNN) is trained using 2D bounding boxes obtained by projecting source-domain 3D annotations onto image planes, rather than nuImages pre-trained weights that may contain target-domain scenes. The 3D detector (ISFusion) is trained on the source split of nuScenes. In \textit{Stage 2}, we freeze the 3D proposal generator and train the fusion decoder for 24 epochs with a batch size of 16 using the AdamW optimizer~\cite{adam}, an initial learning rate of $4\times10^{-4}$, weight decay of $0.01$, and a cosine annealing schedule. Standard data augmentations, including random flipping, rotation, and scaling, are applied during training.

% Paragraph 3: Evaluation Metrics
\smallsection{Evaluation Metrics.}
We report two metrics: (1) \textbf{mAP} (mean Average Precision), averaged over 10 object classes, and (2) \textbf{NDS} (nuScenes Detection Score)~\cite{Caesar2019nuScenesAM}, which combines mAP with translation, scale, orientation, velocity, and attribute errors. Higher values indicate better performance. Owing to the natural data distribution, some target domains do not contain all object categories (\eg, the Night split contains no trailer, construction vehicle, or bus instances). Nevertheless, we report metrics over all 10 classes to remain consistent with the standard nuScenes evaluation protocol.

% Paragraph 4: Baselines and Comparisons
\smallsection{Baselines and Comparisons.}
We compare against representative multi-modal 3D detectors that fuse LiDAR and camera information, including CMT~\cite{yanCrossModalTransformer2023}, MOAD~\cite{chaRobustMultimodal3D2024}, MEFormer~\cite{chaRobustMultimodal3D2024}, ISFusion~\cite{yin2024isfusion}, and MoME~\cite{mome}. For reference, we also include FSDv2~\cite{liFullySparseFusion2024}, a LiDAR-only detector representing single-modality performance. Our method follows a proposal-level fusion paradigm in which modality-specific proposals are generated before cross-modal fusion, with MV2DFusion~\cite{wangMV2DFusionLeveragingModalitySpecific2024} serving as the base framework. For fair comparison, all methods are evaluated under the same data splits, training procedures, and evaluation protocols.

% =============================================
% Subsection 4.2: Main Results
% =============================================
\subsection{Main Results}
\label{sec:exp:main}

% Paragraph 1: Overview of results
% \cref{tab:main_results} presents the cross-domain performance of our method against state-of-the-art baselines. On the source domain, all methods achieve competitive performance, with our approach reaching 68.2\% mAP, demonstrating that our training strategy does not compromise in-domain accuracy. The critical differences emerge on target domains: our method achieves substantial improvements of \textbf{+2.8\%} mAP on Rain, \textbf{+1.3\%} on Night, and \textbf{+3.2\%} on Boston. These gains validate our hypothesis that balanced modality utilization is crucial for robust cross-domain generalization.

\cref{tab:main_results} summarizes the cross-domain detection performance of our method and prior baselines. On the source domain, our method remains competitive, achieving 68.2\% mAP, which indicates that the proposed training strategy preserves in-domain performance. Under domain shift, however, it consistently yields clear gains, improving mAP by \textbf{+2.8\%} on Rain, \textbf{+1.3\%} on Night, and \textbf{+3.2\%} on Boston. These results support our claim that improving modality balance is important for robust cross-domain generalization.

% Paragraph 2: Comparison with multi-modal baselines
\smallsection{Comparison with Multi-Modal Baselines.}
% Feature-level fusion methods show strong source performance but significant target domain degradation: ISFusion~\cite{yin2024isfusion} achieves 66.3\% mAP on source but drops to 39.8/41.8/45.4\% on Rain/Night/Boston, while CMT~\cite{yanCrossModalTransformer2023} and MoME~\cite{mome} reach only 35.7/37.8/42.1\% and 37.7/39.5/42.9\% respectively. This suggests unified query representations lack the structural capacity to adaptively leverage modality-specific information under domain shift. The LiDAR-only FSDv2~\cite{liFullySparseFusion2024} further validates this need, degrading from 59.6\% to 23.4/36.6/28.2\%, highlighting the critical need for robust multi-modal fusion.
Existing multi-modal detectors show reasonable source-domain accuracy but degrade substantially on unseen domains. This trend is particularly evident for methods based on unified feature or query representations, suggesting that they are less effective at preserving modality-specific cues when sensor reliability changes across domains. In contrast, our method is designed to retain modality-specific proposals and strengthen their interaction during fusion, leading to better target-domain robustness.

\oracleresultstable

\begin{figure*}
	\centering
	\includegraphics[width=0.98\linewidth]{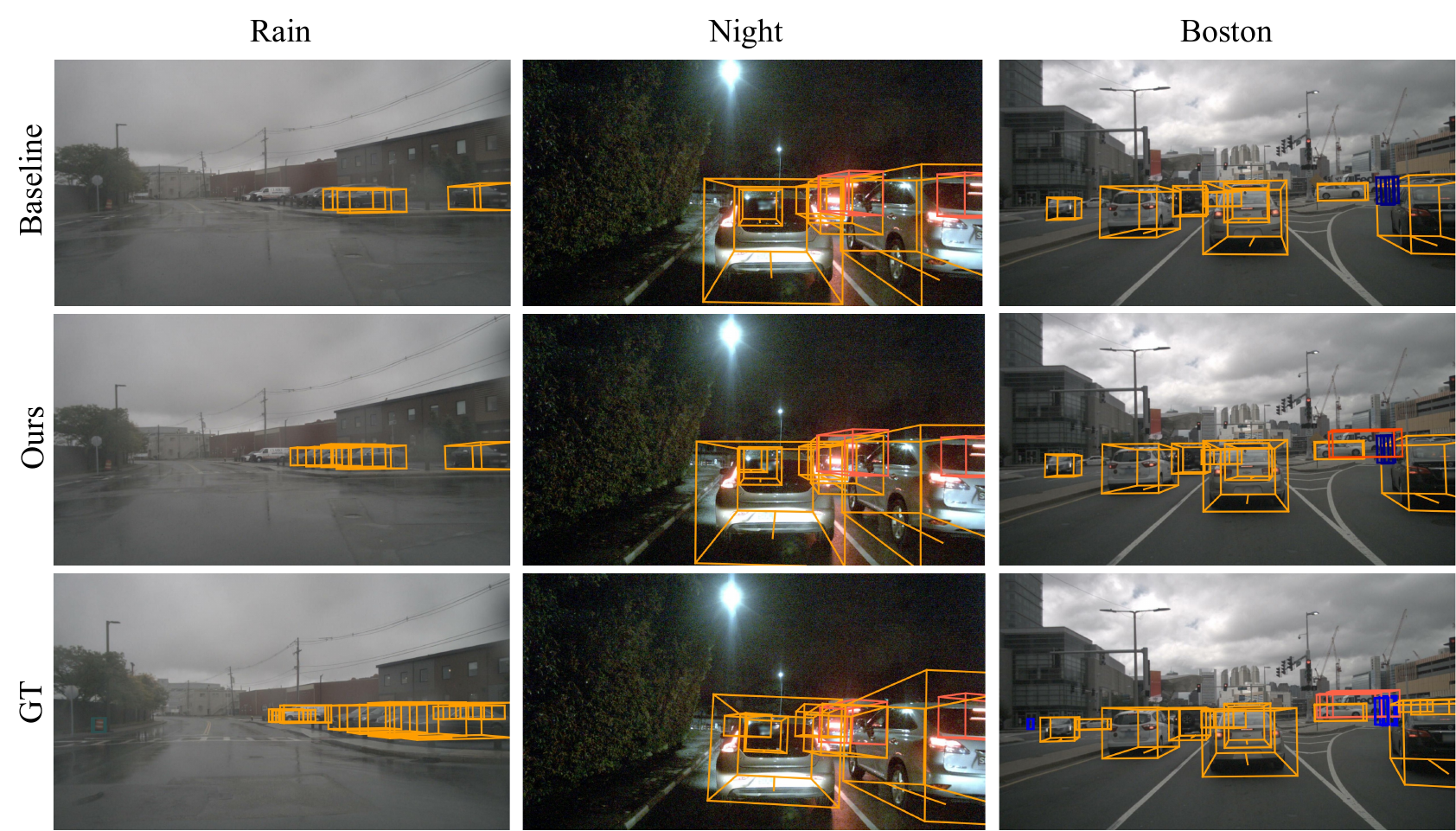}
	\caption{\textbf{Examples of 3D object detections on different data splits. We visualize the 3D bounding boxes of car, truck and pedestrian with \textcolor{orange}{orange}, \textcolor{magenta}{magenta} and \textcolor{blue}{blue} colors in the multi-view images.}}
	\label{fig:case}
\end{figure*}

% Paragraph 3: Our approach and improvements
\smallsection{Our Approach and Improvements.}
Compared with the baseline, our method improves target-domain mAP by 2.8/1.3/3.2 points on Rain/Night/Boston, while maintaining comparable performance on the source domain. These gains come from three complementary components: Query Decoupled Loss strengthens modality-specific supervision, LiDAR-Guided Depth Prior improves the spatial initialization of image queries, and Complementary Cross-Modal Masking promotes more balanced competition between camera- and LiDAR-originated queries. This trend is also reflected in \cref{fig:introduction}(b), where camera-originated queries show substantially larger gains, narrowing the gap to LiDAR-originated queries under domain shift. Beyond mAP, our method also achieves consistently stronger NDS across all domains, indicating improved overall detection quality. Qualitative examples in \cref{fig:case} further show fewer missed detections and false positives.
% under challenging conditions.

% Oracle
To understand the upper bound of performance when trained on target domains, we conduct oracle experiments where models are trained on the standard training split. As shown in \cref{tab:oracle_results}, our method achieves 73.6/72.9/46.9\% mAP on All/Rain/Night splits, outperforming all baselines.

% Paragraph 5: Domain-specific performance
% The performance gains vary across target domains, revealing insights into modality degradation patterns. On Rain, where LiDAR suffers from point sparsity, our method achieves large improvement, demonstrating effective utilization of more reliable image information. On Night, despite challenges in both modalities, our method maintains +1.3\% gain through improved depth estimation. On Boston, the +3.2\% improvement shows that balanced supervision enables robust feature learning under geographic shifts. Beyond mAP, our method consistently outperforms baselines in NDS across all domains (65.9/52.5/45.3/56.8\% vs baseline's 65.7/49.9/44.5/53.6\%), indicating improvements not only in detection recall but also in localization accuracy.

% =============================================
% Subsection 4.3: Ablation Studies
% =============================================
\subsection{Ablation Studies}
\label{sec:exp:ablation}

\ablationtable

% Paragraph 1: Component-wise ablation
We conduct ablation studies to evaluate the contribution of each component. As shown in \cref{tab:ablation}, Query Decoupled Loss improves performance on Rain and Boston, suggesting that stronger modality-specific supervision benefits cross-domain generalization when image cues remain informative. However, it causes a slight drop on Night, where LiDAR is generally more reliable than degraded camera observations. LiDAR-Guided Depth Prior shows a similar trend, as both components primarily strengthen the image branch. In contrast, Complementary Cross-Modal Masking brings consistent gains even without Query Decoupled Loss, indicating that exposing the model to complementary modality degradation is itself effective for improving robustness. When combined with the other components, it further encourages adaptive query selection based on modality reliability rather than fixed preference. Overall, the full model achieves the strongest target-domain mAP performance across the three target splits.

\gridmasktable

% Paragraph 2: Masking strategy ablation
\smallsection{Masking Strategy Design.}
We further analyze the design choices of cross-modal masking in \cref{tab:gridmask}. Starting from the baseline with standard image GridMask~\cite{gridmask}, we compare several masking variants. \textit{Modal Mask}, following CMT~\cite{yanCrossModalTransformer2023} by completely dropping one modality, yields only limited gains because it does not expose the model to complementary partial observations from both modalities. \textit{Consistent GridMask}, which applies the same mask to image and LiDAR inputs, also brings only marginal improvement, as it degrades both modalities simultaneously and thus provides limited support for adaptive fusion. In contrast, \textit{Complementary GridMask} consistently performs better, confirming the importance of preserving complementary observations across modalities. Adding curriculum learning further improves performance, suggesting that progressively increasing masking difficulty stabilizes optimization by allowing the model to first learn from complete inputs and then adapt to partial observations. Replacing GridMask with random masking produces similar results, indicating that the key benefit comes from cross-modal complementarity rather than the specific mask pattern.
\section{Conclusions}
\label{sec:conclusions}

In this work, we study domain generalization for multi-modal 3D object detection in dual-branch proposal-level detectors. We identify modality imbalance as a key limitation of this paradigm: camera-originated queries are often underutilized due to weaker supervision and less reliable spatial initialization, which becomes particularly detrimental under domain shift. To address this issue, we propose Complementary Collaborative Fusion (CCF), a unified framework consisting of Query Decoupled Loss, LiDAR-Guided Depth Prior, and Complementary Cross-Modal Masking. Extensive experiments on challenging real-world domain shifts demonstrate that CCF consistently improves cross-domain robustness over strong baselines while preserving competitive source-domain performance. These results further show the importance of balanced supervision and adaptive query selection for robust multi-modal perception.

\section*{Acknowledgments}
This work was supported in part by the Ministry of Education, Singapore, under its MOE Academic Research Fund Tier 2 (MOE-T2EP20124-0013), and the Agency for Science, Technology and Research (A*STAR) under its MTC Programmatic Funds (Grant No. M23L7b0021). This research work is also supported by Temasek Labs@SUTD.

{
    \small
    \bibliographystyle{ieeenat_fullname}
    \bibliography{main}

@String(CVPR= {IEEE Conf. Comput. Vis. Pattern Recog.})

@String(ICCV= {Int. Conf. Comput. Vis.})

@String(ECCV= {Eur. Conf. Comput. Vis.})

@String(NIPS= {Adv. Neural Inform. Process. Syst.})

@String(ICPR = {Int. Conf. Pattern Recog.})

@String(ICLR = {Int. Conf. Learn. Represent.})

@String(AAAI = {AAAI})

@String(CVPR  = {CVPR})

@String(ICCV  = {ICCV})

@String(ECCV  = {ECCV})

@String(NIPS  = {NeurIPS})

@String(ICPR  = {ICPR})

@String(ICLR  = {ICLR})

@inproceedings{yanCrossModalTransformer2023,
  title = {Cross {{Modal Transformer}}: {{Towards Fast}} and {{Robust 3D Object Detection}}},
  shorttitle = {Cross {{Modal Transformer}}},
  booktitle = ICCV,
  year = {2023},
  author = {Yan, Junjie and Liu, Yingfei and Sun, Jianjian and Jia, Fan and Li, Shuailin and Wang, Tiancai and Zhang, Xiangyu},
  date = {2023-10-01},
  pages = {18222--18232},
  publisher = {IEEE},
  location = {Paris, France},
  doi = {10.1109/ICCV51070.2023.01675},
  url = {https://ieeexplore.ieee.org/document/10377452/},
  urldate = {2024-08-07},
  eventtitle = {2023 {{IEEE}}/{{CVF International Conference}} on {{Computer Vision}} ({{ICCV}})},
  isbn = {979-8-3503-0718-4},
  langid = {english}
}

@article{chaRobustMultimodal3D2024,
      title={Robust Multimodal 3D Object Detection via Modality-Agnostic Decoding and Proximity-based Modality Ensemble}, 
      author={Juhan Cha and Minseok Joo and Jihwan Park and Sanghyeok Lee and Injae Kim and Hyunwoo J. Kim},
      journal={arXiv preprint arXiv:2407.19156},
      year={2024},
}

@inproceedings{yin2024isfusion,
  title={IS-FUSION: Instance-Scene Collaborative Fusion for Multimodal 3D Object Detection},
  author={Yin, Junbo and Shen, Jianbing and Chen, Runnan and Li, Wei and Yang, Ruigang and Frossard, Pascal and Wang, Wenguan},
  booktitle={CVPR},
  year={2024}
}

@inproceedings{mome,
  title={Resilient Sensor Fusion under Adverse Sensor Failures via Multi-Modal Expert Fusion},
  author={Park, Konyul and Kim, Yecheol and Kim, Daehun and Choi, Jun Won},
  booktitle={CVPR},
  year={2025}
}

@inproceedings{liu2022bevfusion,
  title = {{{BEVFusion}}: {{Multi-task}} Multi-Sensor Fusion with Unified Bird's-Eye View Representation},
  booktitle = {{{IEEE}} International Conference on Robotics and Automation ({{ICRA}})},
  author = {Liu, Zhijian and Tang, Haotian and Amini, Alexander and Yang, Xingyu and Mao, Huizi and Rus, Daniela and Han, Song},
  year = {2023},
}

@ARTICLE{wangMV2DFusionLeveragingModalitySpecific2024,
  author={Wang, Zitian and Huang, Zehao and Gao, Yulu and Wang, Naiyan and Liu, Si},
  journal={IEEE Transactions on Pattern Analysis and Machine Intelligence}, 
  title={MV2DFusion: Leveraging Modality-Specific Object Semantics for Multi-Modal 3D Detection}, 
  year={2025},
  volume={},
  number={},
  pages={1-15},
  doi={10.1109/TPAMI.2025.3609348}
}

@article{liFullySparseFusion2024,
  title = {Fully {{Sparse Fusion}} for {{3D Object Detection}}},
  author = {Li, Yingyan and Fan, Lue and Liu, Yang and Huang, Zehao and Chen, Yuntao and Wang, Naiyan and Zhang, Zhaoxiang},
  year = {2024},
  journal = {IEEE Transactions on Pattern Analysis and Machine Intelligence},
  shortjournal = {IEEE Trans. Pattern Anal. Mach. Intell.},
  eprint = {2304.12310},
  eprinttype = {arXiv},
  eprintclass = {cs},
  pages = {1--15},
  issn = {0162-8828, 2160-9292, 1939-3539},
  doi = {10.1109/TPAMI.2024.3392303},
  url = {http://arxiv.org/abs/2304.12310},
  urldate = {2024-07-04},
  langid = {english}
}

@INPROCEEDINGS{xieSparseFusionFusingMultiModal2023,
  author={Xie, Yichen and Xu, Chenfeng and Rakotosaona, Marie-Julie and Rim, Patrick and Tombari, Federico and Keutzer, Kurt and Tomizuka, Masayoshi and Zhan, Wei},
  booktitle={2023 IEEE/CVF International Conference on Computer Vision (ICCV)}, 
  title={SparseFusion: Fusing Multi-Modal Sparse Representations for Multi-Sensor 3D Object Detection}, 
  year={2023},
  volume={},
  number={},
  pages={17545-17556},
  doi={10.1109/ICCV51070.2023.01613}}

@article{Caesar2019nuScenesAM,
  title = {{{nuScenes}}: A Multimodal Dataset for Autonomous Driving},
  author = {Caesar, Holger and Bankiti, Varun and Lang, Alex H. and Vora, Sourabh and Liong, Venice Erin and Xu, Qiang and Krishnan, Anush and Pan, Yuxin and Baldan, Giancarlo and Beijbom, Oscar},
  year = {2019},
  journal = {2020 IEEE/CVF Conference on Computer Vision and Pattern Recognition (CVPR)},
  pages = {11618--11628},
  url = {https://api.semanticscholar.org/CorpusID:85517967}
}

@article{huangBEVDetHighperformanceMulticamera2022,
  title = {{{BEVDet}}: {{High-performance Multi-camera 3D Object Detection}} in {{Bird-Eye-View}}},
  shorttitle = {{{BEVDet}}},
  author = {Huang, Junjie and Huang, Guan and Zhu, Zheng and Ye, Yun and Du, Dalong},
  year = {2022},
  journal={arXiv preprint arXiv:2112.11790},
  eprint = {2112.11790},
  eprinttype = {arXiv},
  eprintclass = {cs},
  url = {http://arxiv.org/abs/2112.11790},
  urldate = {2022-07-12},
  pubstate = {prepublished},
}

@inproceedings{PETR,
  title = {{{PETR}}: Position Embedding Transformation for Multi-View {{3D}} Object Detection},
  booktitle = {Computer Vision - {{ECCV}} 2022 - 17th European Conference, Tel Aviv, Israel, October 23-27, 2022, Proceedings, Part {{XXVII}}},
  author = {Liu, Yingfei and Wang, Tiancai and Zhang, Xiangyu and Sun, Jian},
  editor = {Avidan, Shai and Brostow, Gabriel J. and Cissé, Moustapha and Farinella, Giovanni Maria and Hassner, Tal},
  year = {2022},
  series = {Lecture Notes in Computer Science},
  volume = {13687},
  pages = {531--548},
  publisher = {Springer},
  doi = {10.1007/978-3-031-19812-0\_31},
  url = {https://doi.org/10.1007/978-3-031-19812-0_31},
  bibsource = {dblp computer science bibliography, https://dblp.org}
}

@inproceedings{centerpoint,
  title = {Center-Based {{3D}} Object Detection and Tracking},
  booktitle = {{{IEEE}} Conference on Computer Vision and Pattern Recognition, {{CVPR}} 2021, Virtual, June 19-25, 2021},
  author = {Yin, Tianwei and Zhou, Xingyi and Krähenbühl, Philipp},
  year = {2021},
  pages = {11784--11793},
  publisher = {Computer Vision Foundation / IEEE},
  doi = {10.1109/CVPR46437.2021.01161},
  url = {https://openaccess.thecvf.com/content/CVPR2021/html/Yin_Center-Based_3D_Object_Detection_and_Tracking_CVPR_2021_paper.html},
  bibsource = {dblp computer science bibliography, https://dblp.org}
}

@inproceedings{Transfusion,
  title = {{{TransFusion}}: {{Robust LiDAR-camera}} Fusion for {{3D}} Object Detection with Transformers},
  booktitle = {{{IEEE}}/{{CVF}} Conference on Computer Vision and Pattern Recognition, {{CVPR}} 2022, New Orleans, {{LA}}, {{USA}}, June 18-24, 2022},
  author = {Bai, Xuyang and Hu, Zeyu and Zhu, Xinge and Huang, Qingqiu and Chen, Yilun and Fu, Hongbo and Tai, Chiew-Lan},
  year = {2022},
  pages = {1080--1089},
  publisher = {IEEE},
  doi = {10.1109/CVPR52688.2022.00116},
  url = {https://doi.org/10.1109/CVPR52688.2022.00116},
  bibsource = {dblp computer science bibliography, https://dblp.org}
}

@inproceedings{GraphBEV,
  title = {{{GraphBEV}}: {{Towards}} Robust {{BEV}} Feature Alignment for Multi-Modal {{3D}} Object Detection},
  booktitle = {Computer Vision - {{ECCV}} 2024 - 18th European Conference, Milan, Italy, September 29-October 4, 2024, Proceedings, Part {{XXVI}}},
  author = {Song, Ziying and Yang, Lei and Xu, Shaoqing and Liu, Lin and Xu, Dongyang and Jia, Caiyan and Jia, Feiyang and Wang, Li},
  editor = {Leonardis, Ales and Ricci, Elisa and Roth, Stefan and Russakovsky, Olga and Sattler, Torsten and Varol, Gül},
  year = {2024},
  series = {Lecture Notes in Computer Science},
  volume = {15084},
  pages = {347--366},
  publisher = {Springer},
  doi = {10.1007/978-3-031-73347-5\_20},
  url = {https://doi.org/10.1007/978-3-031-73347-5_20},
  bibsource = {dblp computer science bibliography, https://dblp.org}
}

@article{fasterrcnn,
  title={Faster R-CNN: Towards Real-Time Object Detection with Region Proposal Networks},
  author={Shaoqing Ren and Kaiming He and Ross B. Girshick and Jian Sun},
  journal={IEEE Transactions on Pattern Analysis and Machine Intelligence},
  year={2015},
  volume={39},
  pages={1137-1149},
  url={https://api.semanticscholar.org/CorpusID:10328909}
}

@article{focalloss,
  title={Focal Loss for Dense Object Detection},
  author={Tsung-Yi Lin and Priya Goyal and Ross B. Girshick and Kaiming He and Piotr Doll{\'a}r},
  journal={2017 IEEE International Conference on Computer Vision (ICCV)},
  year={2017},
  pages={2999-3007},
  url={https://api.semanticscholar.org/CorpusID:47252984}
}

@inbook{poe,
  author = {G.E. Hinton },
  title = {Products of experts},
  booktitle = {ICANN99. Ninth International Conference on Artificial Neural Networks (IEE Conf. Publ. No.470)},
  year = {1999},
  publisher = {IEEE},
  chapter = {},
  pages = {1-6},
  doi = {10.1049/cp:19991075},
  URL = {https://digital-library.theiet.org/doi/abs/10.1049/cp%3A19991075},
  eprint = {https://digital-library.theiet.org/doi/pdf/10.1049/cp%3A19991075},
}

@article{gridmask,
  author       = {Pengguang Chen and Shu Liu and Hengshuang Zhao and Jiaya Jia},
  title        = {GridMask Data Augmentation},
  journal      = {arXiv preprint arXiv:2001.04086},
  year         = {2020},
}

@inproceedings{adam,
  author       = {Diederik P. Kingma and
                  Jimmy Ba},
  editor       = {Yoshua Bengio and
                  Yann LeCun},
  title        = {Adam: {A} Method for Stochastic Optimization},
  booktitle    = {3rd International Conference on Learning Representations, {ICLR} 2015,
                  San Diego, CA, USA, May 7-9, 2015, Conference Track Proceedings},
  year         = {2015},
  url          = {http://arxiv.org/abs/1412.6980},
  bibsource    = {dblp computer science bibliography, https://dblp.org}
}

@inproceedings{metabev2025,
	author={Ge, Chongjian and Chen, Junsong and Xie, Enze and Wang, Zhongdao and Hong, Lanqing and Lu, Huchuan and Li, Zhenguo and Luo, Ping},
	booktitle=ICCV, 
	title={MetaBEV: Solving Sensor Failures for 3D Detection and Map Segmentation}, 
	year={2023}}

@inproceedings{wang2023unibev,
	title={UniBEV: Multi-modal 3D Object Detection with Uniform BEV Encoders for Robustness against Missing Sensor Modalities},
	author={Wang, Shiming and Caesar, Holger and Nan, Liangliang and Kooij, Julian FP},
	booktitle={2024 IEEE Intelligent Vehicles Symposium (IV)},
	year={2024}
}

@inproceedings{xiao2022polarmix,
author = {Xiao, Aoran and Huang, Jiaxing and Guan, Dayan and Cui, Kaiwen and Lu, Shijian and Shao, Ling},
title = {PolarMix: a general data augmentation technique for LiDAR point clouds},
year = {2022},
isbn = {9781713871088},
publisher = {Curran Associates Inc.},
address = {Red Hook, NY, USA},
booktitle = {Proceedings of the 36th International Conference on Neural Information Processing Systems},
articleno = {802},
numpages = {14},
location = {New Orleans, LA, USA},
series = {NIPS '22}
}

@inproceedings{kong2023lasermix,
	title = {LaserMix for Semi-Supervised LiDAR Semantic Segmentation},
	author = {Kong, Lingdong and Ren, Jiawei and Pan, Liang and Liu, Ziwei},
	booktitle = {IEEE/CVF Conference on Computer Vision and Pattern Recognition},
	pages = {21705--21715},
	year = {2023},
}

@article{park2024rethinking,
	title={Rethinking Data Augmentation for Robust LiDAR Semantic Segmentation in Adverse Weather},
	author={Park, Junsung and Kim, Kyungmin and Shim, Hyunjung},
	journal={arXiv preprint arXiv:2407.02286},
	year={2024}
}

@article{fpnet,
  title={Frustum PointNets for 3D Object Detection from RGB-D Data},
  author={C. Qi and W. Liu and Chenxia Wu and Hao Su and Leonidas J. Guibas},
  journal={2018 IEEE/CVF Conference on Computer Vision and Pattern Recognition},
  year={2017},
  pages={918-927},
  url={https://api.semanticscholar.org/CorpusID:4868248}
}

@INPROCEEDINGS{objfusion,
  author={Cai, Qi and Pan, Yingwei and Yao, Ting and Ngo, Chong-Wah and Mei, Tao},
  booktitle={2023 IEEE/CVF International Conference on Computer Vision (ICCV)}, 
  title={ObjectFusion: Multi-modal 3D Object Detection with Object-Centric Fusion}, 
  year={2023},
  volume={},
  number={},
  pages={18021-18030},
  doi={10.1109/ICCV51070.2023.01656}}

@ARTICLE{cofix,
  author={Li, Wenxuan and Zou, Qin and Chen, Chi and Du, Bo and Chen, Long and Zhou, Jian and Yu, Hongkai},
  journal={IEEE Robotics and Automation Letters}, 
  title={Co-Fix3D: Enhancing 3D Object Detection With Collaborative Refinement}, 
  year={2025},
  volume={10},
  number={5},
  pages={4970-4977},
  doi={10.1109/LRA.2025.3555859}}

@ARTICLE{bico,
  author={Song, Yang and Wang, Lin},
  journal={IEEE Robotics and Automation Letters}, 
  title={BiCo-Fusion: Bidirectional Complementary LiDAR-Camera Fusion for Semantic- and Spatial-Aware 3D Object Detection}, 
  year={2025},
  volume={10},
  number={2},
  pages={1457-1464},
  doi={10.1109/LRA.2024.3518845}}

@INPROCEEDINGS{mv3d,
  author={Chen, Xiaozhi and Ma, Huimin and Wan, Ji and Li, Bo and Xia, Tian},
  booktitle={2017 IEEE Conference on Computer Vision and Pattern Recognition (CVPR)}, 
  title={Multi-view 3D Object Detection Network for Autonomous Driving}, 
  year={2017},
  volume={},
  number={},
  pages={6526-6534},
  doi={10.1109/CVPR.2017.691}}

@INPROCEEDINGS{paint,
  author={Xu, Shaoqing and Zhou, Dingfu and Fang, Jin and Yin, Junbo and Bin, Zhou and Zhang, Liangjun},
  booktitle={2021 IEEE International Intelligent Transportation Systems Conference (ITSC)}, 
  title={FusionPainting: Multimodal Fusion with Adaptive Attention for 3D Object Detection}, 
  year={2021},
  volume={},
  number={},
  pages={3047-3054},
  doi={10.1109/ITSC48978.2021.9564951}}

@inproceedings{IAL,
  author       = {Yining Pan and
                  Qiongjie Cui and
                  Xulei Yang and
                  Na Zhao},
  editor       = {Aarti Singh and
                  Maryam Fazel and
                  Daniel Hsu and
                  Simon Lacoste{-}Julien and
                  Felix Berkenkamp and
                  Tegan Maharaj and
                  Kiri Wagstaff and
                  Jerry Zhu},
  title        = {How Do Images Align and Complement LiDAR? Towards a Harmonized Multi-modal
                  3D Panoptic Segmentation},
  booktitle    = {Forty-second International Conference on Machine Learning, {ICML}
                  2025, Vancouver, BC, Canada, July 13-19, 2025},
  series       = {Proceedings of Machine Learning Research},
  volume       = {267},
  publisher    = {{PMLR} / OpenReview.net},
  year         = {2025},
  url          = {https://proceedings.mlr.press/v267/pan25c.html},
  bibsource    = {dblp computer science bibliography, https://dblp.org}
}

@INPROCEEDINGS{panopticph,
  author={Li, Jinke and He, Xiao and Wen, Yang and Gao, Yuan and Cheng, Xiaoqiang and Zhang, Dan},
  booktitle={2022 IEEE/CVF Conference on Computer Vision and Pattern Recognition (CVPR)}, 
  title={Panoptic-PHNet: Towards Real-Time and High-Precision LiDAR Panoptic Segmentation via Clustering Pseudo Heatmap}, 
  year={2022},
  volume={},
  number={},
  pages={11799-11808},
  doi={10.1109/CVPR52688.2022.01151}}

@inproceedings{spgroup3d,
  title={Spgroup3d: Superpoint grouping network for indoor 3d object detection},
  author={Zhu, Yun and Hui, Le and Shen, Yaqi and Xie, Jin},
  booktitle={Proceedings of the AAAI Conference on Artificial Intelligence},
  volume={38},
  number={7},
  pages={7811--7819},
  year={2024}
}

@InProceedings{Zhu_2025_CVPR,
    author    = {Zhu, Yun and Hui, Le and Yang, Hang and Qian, Jianjun and Xie, Jin and Yang, Jian},
    title     = {Learning Class Prototypes for Unified Sparse-Supervised 3D Object Detection},
    booktitle = {Proceedings of the IEEE/CVF Conference on Computer Vision and Pattern Recognition (CVPR)},
    month     = {June},
    year      = {2025},
    pages     = {9911-9920}
}

@inproceedings{zhao2021ps2,
  title={Psˆ2-net: A locally and globally aware network for point-based semantic segmentation},
  author={Zhao, Na and Chua, Tat-Seng and Lee, Gim Hee},
  booktitle={2020 25th International Conference on Pattern Recognition (ICPR)},
  pages={723--730},
  year={2021},
  organization={IEEE}
}

@inproceedings{zhao2020sess,
  title={Sess: Self-ensembling semi-supervised 3d object detection},
  author={Zhao, Na and Chua, Tat-Seng and Lee, Gim Hee},
  booktitle={Proceedings of the IEEE/CVF Conference on Computer Vision and Pattern Recognition},
  pages={11079--11087},
  year={2020}
}

@inproceedings{zhao2021few,
  title={Few-shot 3d point cloud semantic segmentation},
  author={Zhao, Na and Chua, Tat-Seng and Lee, Gim Hee},
  booktitle={Proceedings of the IEEE/CVF conference on computer vision and pattern recognition},
  pages={8873--8882},
  year={2021}
}

@inproceedings{sheng2022rethinking,
  title={Rethinking IoU-based optimization for single-stage 3D object detection},
  author={Sheng, Hualian and Cai, Sijia and Zhao, Na and Deng, Bing and Huang, Jianqiang and Hua, Xian-Sheng and Zhao, Min-Jian and Lee, Gim Hee},
  booktitle={European Conference on Computer Vision},
  pages={544--561},
  year={2022},
  organization={Springer}
}

@inproceedings{xu2023generalized,
  title={Generalized few-shot point cloud segmentation via geometric words},
  author={Xu, Yating and Hu, Conghui and Zhao, Na and Lee, Gim Hee},
  booktitle={Proceedings of the IEEE/CVF International Conference on Computer Vision},
  pages={21506--21515},
  year={2023}
}

@inproceedings{zhao2024synthetic,
  title={Synthetic-to-Real Domain Generalized Semantic Segmentation for 3D Indoor Point Clouds},
  author={Zhao, Yuyang and Zhao, Na and Lee, Gim Hee},
  booktitle={The British Machine Vision Conference},
  year={2024}
}

@inproceedings{han2024dual,
  title={Dual-perspective knowledge enrichment for semi-supervised 3d object detection},
  author={Han, Yucheng and Zhao, Na and Chen, Weiling and Ma, Keng Teck and Zhang, Hanwang},
  booktitle={Proceedings of the AAAI Conference on Artificial Intelligence},
  volume={38},
  number={3},
  pages={2049--2057},
  year={2024}
}

@article{zhao2024sdcot++,
  title={SDCoT++: Improved Static-Dynamic Co-Teaching for Class-Incremental 3D Object Detection},
  author={Zhao, Na and Qian, Peisheng and Wu, Fang and Xu, Xun and Yang, Xulei and Lee, Gim Hee},
  journal={IEEE Transactions on Image Processing},
  volume={34},
  pages={4188--4202},
  year={2024},
  publisher={IEEE}
}

@article{sheng2025ct3d++,
  title={Ct3d++: Improving 3d object detection with keypoint-induced channel-wise transformer},
  author={Sheng, Hualian and Cai, Sijia and Zhao, Na and Deng, Bing and Liang, Qiao and Zhao, Min-Jian and Ye, Jieping},
  journal={International Journal of Computer Vision},
  volume={133},
  number={7},
  pages={4817--4836},
  year={2025},
  publisher={Springer}
}

@inproceedings{wang2025uncertainty,
  title={Uncertainty meets diversity: A comprehensive active learning framework for indoor 3d object detection},
  author={Wang, Jiangyi and Zhao, Na},
  booktitle={Proceedings of the Computer Vision and Pattern Recognition Conference},
  pages={20329--20339},
  year={2025}
}
}

% \appendix

\end{document}